# Land cover and flood type govern the detection limits of satellite-based flood mapping across diverse global flood events

Venkatesh Kolluru[1,*], Rajat Shinde[1], Abdelhak Marouane[1], Caden Helbling[1]; Deepak Shah[1]; Othneil Drew[2]; Iksha Gurung[1]; Manil Maskey[3]; Rahul Ramachandran[3]

[1]Earth System Science Center, University of Alabama in Huntsville, Huntsville, AL 35805, USA
[2]Space and Earth Science Data Analysis, 7515 Mission Drive, Lanham, MD 20706
[3]NASA Marshall Space Flight Center, Huntsville, AL, 35808 USA.

*Corresponding author
Venkatesh Kolluru
Earth System Science Center, University of Alabama in Huntsville, Huntsville, AL 35805, USA
Email: vk0046@uah.edu
ORCID: 0000-0002-2110-5560

## Abstract

Floods are among the most destructive natural hazards, and their increasing frequency under climate change makes satellite-based inundation mapping essential for disaster response. Geospatial foundation models pretrained on satellite archives offer geographic transferability, but their operational reliability across diverse, unseen events remains uncharacterized. Here we deploy Prithvi-EO-2.0 across 19 out-of-distribution flood events (2017-2025) spanning six continents, eight climate zones, and six flood mechanisms, validating against two independent reference products. Detection accuracy depended jointly on land cover and flood type, with cropland yielding the highest agreement (IoU=52%) and riverine events the strongest detection (F1=0.69), while tree cover and built-up areas showed near-zero detection (IoU=4%) regardless of flood mechanism. Dual-reference validation revealed that apparent model error partly reflects definitional inconsistency between reference products rather than detection failure. Iterative pipeline testing identified 23 failure modes, with pipeline engineering dominating initial error over model capacity. These findings establish environment-dependent detection boundaries for operational satellite flood mapping.

Keywords: Climate Change; Disaster events; Failure modes; Foundation model; HLS; Red teaming; Segmentation

# Introduction

Floods account for over 40% of global disasters and remain the most frequent and devastating weather-related natural hazard worldwide[1]. Between 1990 and 2022, 4713 flood events across 168 countries affected over 3.2 billion people, caused ~0.21M deaths, and resulted in economic losses exceeding $1.3 trillion[2,3]. These economic ramifications and the increasing frequency of flood disasters underscore the need for accurate, timely satellite-based flood-extent mapping to support emergency response, recovery, and risk mitigation[4,5].

Satellite remote sensing underpins large-scale flood mapping, and the availability of multisensor composites with a 2-to-3-day global revisit has enabled consistent flood monitoring capability[6,7,8] and increased the coverage of viable flood imagery from 7.2% pre-2014 to 63.9% post-2014[9]. However, leveraging this growing data volume for operational mapping requires automated processing beyond existing workflows. In recent years, deep learning models such as U-Net have extracted hierarchical spatial features without manual feature design[10,11]. However, these models failed to generalize across regions, constrained by both the limited availability of high-quality labeled flood training data[12] and the domain specificity of existing benchmarks. Moreover, these models achieving high accuracy on individual benchmarks do not consistently transfer to other datasets[13,14]. This generalization gap, coupled with the cost of generating labeled data, has motivated the development of geospatial foundation models (GeoFMs) trained via self-supervised learning over large volumes of unlabeled satellite imagery[15,16].

Among them, Prithvi-EO-2.0, developed by NASA and IBM, a 600-million-parameter temporal ViT pretrained on over 4.2 million global HLS chips and fine-tuned on Sen1Floods11, has shown strong performance for flood inundation mapping[17,18,19]. Studies demonstrate that the pretrained Prithvi ViT encoder paired with a U-Net decoder outperforms a U-Net trained from scratch in out-of-distribution and few-shot settings due to improved geographic transferability, while hybrid approaches such as U-Prithvi balance in-sample and out-of-distribution (OOD) accuracy[20,21,22]. However, GeoFM development has substantially outpaced task-specific operational deployment[23,24]. The engineering complexity of integrating pretrained models into end-to-end processing pipelines spanning data acquisition, preprocessing, inference, postprocessing, and validation has been the primary barrier limiting GeoFM adoption in operational settings. Moreover, challenges such as input specification constraints, definitional gaps between pretraining objectives and operational requirements, and pipeline failure propagation across these stages remain uncharacterized. Consequently, existing flood mapping evaluations are conducted under controlled benchmark conditions with OOD assessment restricted to one or two held-out regions and validation against a single reference product. For example, Li et al. (2023)[20] evaluated Prithvi on a single OOD event (Bolivia), Kaushik et al. (2026)[22] used leave-one-region-out with five sites, and Tulbure et al. (2025)[25] validated against OPERA DSWx alone. This leaves the question of GeoFM operational reliability across diverse, globally distributed events largely unanswered.

Currently operational deep learning systems highlight both the feasibility and challenges that emerge beyond benchmark settings, including false positive accumulation[26], inflated accuracy from subjective evaluation choices[27], and definitional differences between reference products[28]. Additionally, operational products such as Copernicus Global Flood Monitoring (GFM) and the OPERA Dynamic Surface Water Extent (DSWx-HLS) differ in their definitions of flood versus surface water, complicating cross-validation[28]. Collectively, these systems indicate that operational deployment introduces data-handling, preprocessing, and validation challenges absent from benchmark settings. However, no study has systematically documented or categorized these pipeline failure modes. Red teaming, a structured adversarial AI evaluation methodology[29], provides such a framework for identifying vulnerabilities overlooked by conventional validation[29,30]. In the context of geospatial AI, red teaming can expose data handling inconsistencies, inference artifacts, definitional mismatches between model outputs and reference products, and cascading errors across pipeline stages that degrade end-to-end reliability. To our knowledge, no published study has applied red teaming or systematic failure mode analysis to GeoFM deployment for flood mapping.

Moreover, existing GeoFM-based flood evaluations report aggregate metrics (e.g., IoU and F1) at the scene or event level[20,22,25], which obscure context-dependent failure modes critical to operational reliability. Studies have also shown that land cover, topography, and climate strongly influence flood characteristics and model performance[31,32]. However, GeoFM evaluations have not stratified detection accuracy by these environmental variables, limiting understanding of context-dependent reliability. Similarly, physics-based effects such as terrain shadows, atmospheric aerosols, and vegetation-induced false positives are known to affect optical flood classification[33,34]. However, their quantitative impact on foundation model outputs has not been isolated from model accuracy. Furthermore, flood extents vary widely in spatial scale, from sub-kilometer fragments to contiguous regions spanning thousands of square kilometers. However, patch-level spatial analyses that quantify detection sensitivity and fragmentation remain largely absent from the GeoFM evaluation literature, obscuring the minimum detectable flood extent for disaster response. Together, these gaps indicate that current evaluations measure model performance in isolation, rather than within the environmental and system-level contexts that govern operational effectiveness.

This study addresses these gaps by deploying Prithvi-EO-2.0 for flood mapping across 19 globally distributed OOD events (2017-2025), spanning six continents, multiple climate zones, and diverse flood mechanisms, with no geographic or temporal overlap with the Sen1Floods11 training data (Fig. 1). We apply a red-teaming methodology to systematically identify and categorize failure modes across the complete processing pipeline, from data acquisition through inference to validation, producing the first operational failure mode taxonomy for GeoFM-based flood mapping. Predictions are validated against the Copernicus GFM (n = 19 events) and OPERA DSWx-HLS (n = 12 events), providing dual-reference cross-validation that accounts for definitional differences between flood water and surface water detection. We quantify the impact of physics-based postprocessing (terrain shadow correction, aerosol filtering, vegetation removal, permanent water masking) on prediction accuracy and stratify performance by flood size, Köppen climate zone, flood mechanism, and ESA WorldCover land cover class. We conduct patch-level analysis across 79,885 patches to characterize detection recall as a function of patch size (Fig. 2). Specifically, this study addresses five research questions: (1) How accurately does Prithvi-EO-2.0 detect floods in OOD events across diverse geographic and climatic contexts? (2) What systematic failure modes emerge during end-to-end operational deployment, and how do pipeline stages contribute to interruptions? (3) How does detection accuracy vary with flood size, climate zone, flood mechanism, and land cover class? (4) What is the quantitative impact of physics-based postprocessing on flood extent estimates and accuracy? (5) How do predictions validated against two independent reference products compare in precision, recall, and agreement patterns?

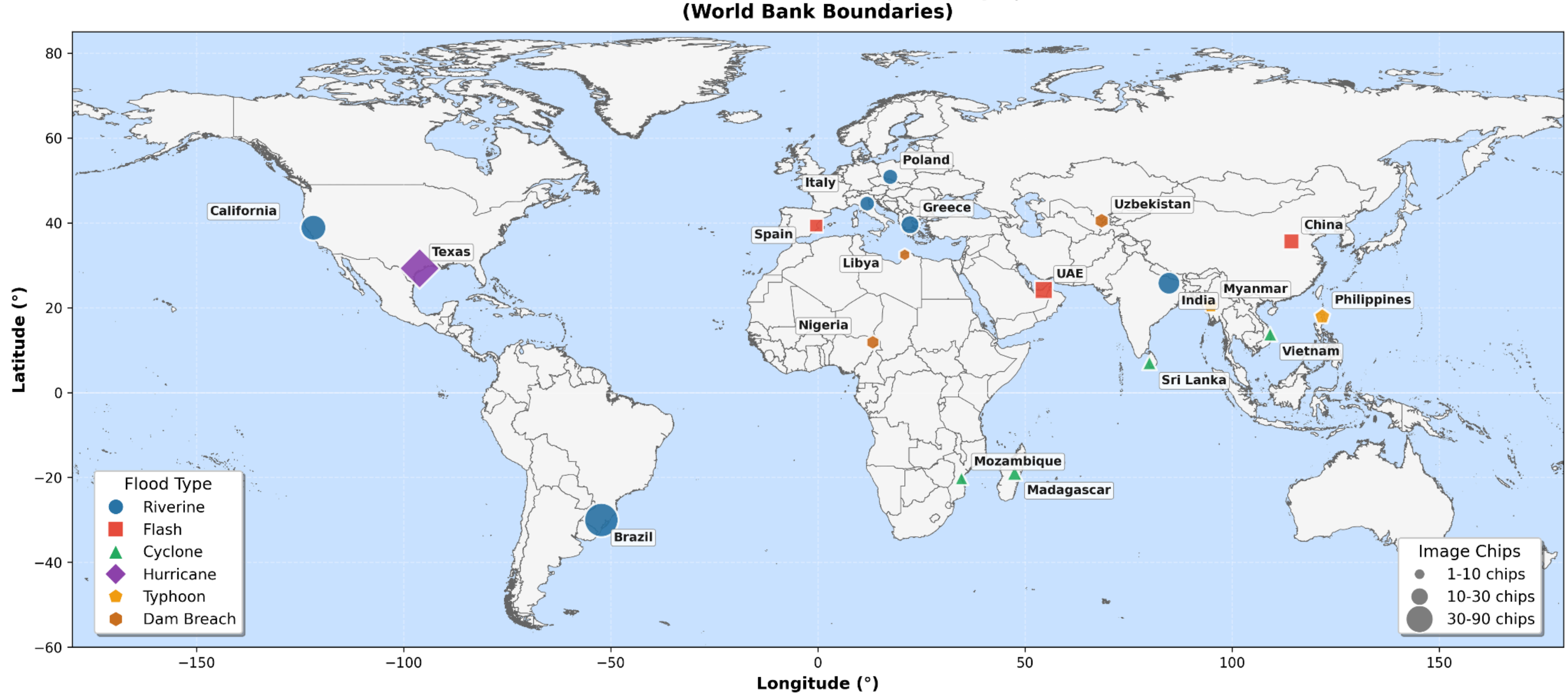

***Figure 1.*** *Global distribution of 19 flood events used for model evaluation. Marker shapes indicate flood type, and marker size is scaled proportionally to the number of image chips per event across three classes: small (1–10 chips), medium (10–30 chips), and large (30–90 chips). Country boundaries are from the World Bank Official Boundaries dataset. Chips denote the number of non-overlapping 512 x 512 pixel image tiles per event.*

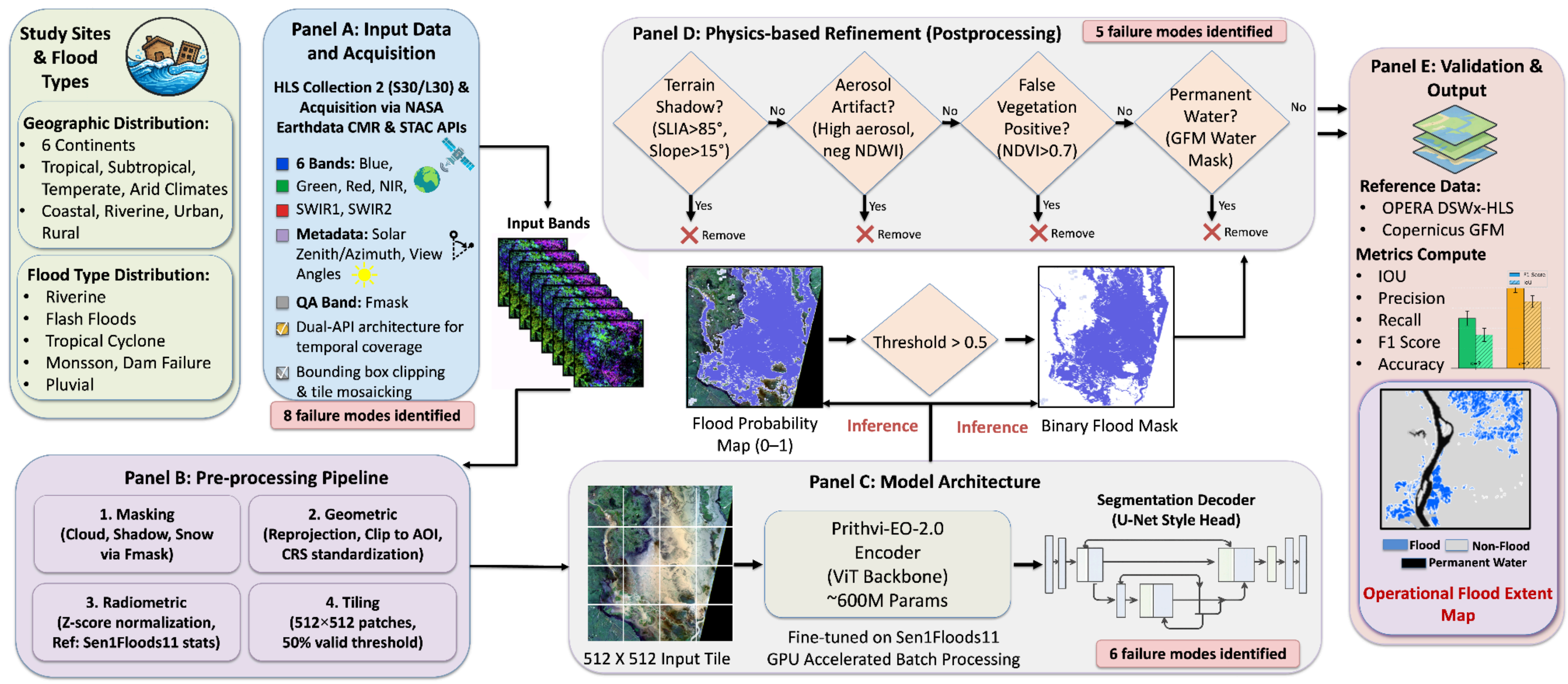

***Figure 2.*** *End-to-end flood mapping methodology using the Prithvi-EO-2.0 foundation model. Panel A: HLS data acquisition via NASA Earthdata CMR/STAC APIs. Panel B: Preprocessing (cloud masking, geometric correction, radiometric normalization, tiling). Panel C: ViT-based encoder (~600M parameters) with U-Net decoder producing binary flood masks at threshold 0.5. Panel D: Physics-based postprocessing filters for terrain shadow, aerosol artifacts, vegetation false positives, and permanent water. Panel E: Validation against OPERA DSWx-HLS and Copernicus GFM. Failure mode counts are annotated at each pipeline stage.*

Our findings provide practical guidance for researchers and practitioners seeking to operationalize GeoFMs for flood mapping, including which (a) preprocessing and postprocessing strategies to implement, (b) systematic failure modes to anticipate during pipeline development, and (c) validation practices to adopt when deploying models to out-of-distribution events. To our knowledge, this is the first study to apply a red-teaming framework to the operational deployment of a geospatial foundation model, test model performance across 19 globally distributed OOD flood events, and simultaneously validate flood predictions against two independent reference products. We anticipate that as foundation models evolve, the evaluation methodology, failure mode taxonomy, and dual-reference validation framework presented here will apply to future GeoFM assessments for flood mapping and other Earth observation tasks.

## Results

### Per-Event Flood Detection Performance

We tested the robustness of Prithvi-EO-2.0 in detecting floods across 19 OOD flood events and mapped a total flood extent of 10,578.6 km² spanning six continents, eight climate zones, and six flood mechanisms. We then compared our model predictions against two reference products (GFM and DSWx) using six standard accuracy metrics (Table 3, Fig. 3). We obtained a mean accuracy of 89.94% (ranging from 76-96% across 19 events) and 89.78% (ranging from 82-93% across 12 events), when compared against GFM and DSWx, respectively. The F1 scores against GFM ranged from 0.24 (Myanmar) to 0.88 (Libya), with a mean and median of 0.61 and 0.60, respectively. The IoU ranged from 0.13 (Myanmar) to 0.79 (Libya) with a mean of 0.46. The model validation against GFM showed high recall (mean: 0.90, median: 0.93), low precision (mean: 0.49, median: 0.48), and greater bias (>1.0 for all 19 events), indicating systematic overdetection in predictions. We observed a contrasting pattern in accuracy against DSWx. The model yielded a mean F1 of 0.65 (range: 0.44 to 0.83, n=12) and a mean IoU of 0.49 (range: 0.28 to 0.71). DSWx validation yielded higher precision (mean: 0.75) but lower recall (mean: 0.60), with frequency bias below 1.0 for 9 of 12 events (mean: 0.82). This contrasting pattern reflects the definitional difference between the two references: GFM maps only transient floodwater (excluding permanent water bodies), while DSWx maps total surface water, including permanent water bodies. Additional detail on the definitional differences and their implications is provided in Supplementary Text S1.

Overall, the per-event analysis demonstrates that Prithvi-EO-2.0 achieves operationally useful flood detection (F1 > 0.7) for 10 of 19 OOD events without fine-tuning, but exhibits substantial variability driven by land cover complexity, cloud contamination, and differences in reference product definitions. The systematic overdetection against GFM and underdetection against DSWx positions the model predictions between the flood-only and total-water reference definitions.

### Impact of Physics-Based Postprocessing

We reported the flood area before and after correction for each event to isolate the contribution of the four-stage postprocessing (terrain shadow removal, aerosol filtering, vegetation masking, and permanent water correction). We found a total area reduction of 277.41 km² across 19 events (Table 4), with event-level reductions ranging from 0% (Libya, Spain) to 18.93% (Italy). Four events exhibited reductions exceeding 5%: Italy (18.93%), Madagascar (7.67%), China (6.14%), and Texas (5.86%). These events share terrain characteristics susceptible to shadow false positives, namely, the Apennine foothills (Italy), central highlands (Madagascar), mountainous terrain adjacent to the Henan floodplain (China), and coastal topographic variation (Texas). Although the aggregate correction is modest, the postprocessing disproportionately affects events with complex terrain where uncorrected shadow artifacts would systematically inflate flood extent estimates. Events in flat or arid terrain (Libya, Spain, UAE) showed minimal or zero reduction, consistent with the absence of topographic shadow sources. The mean area reduction was 3.04% (median: 1.78%).

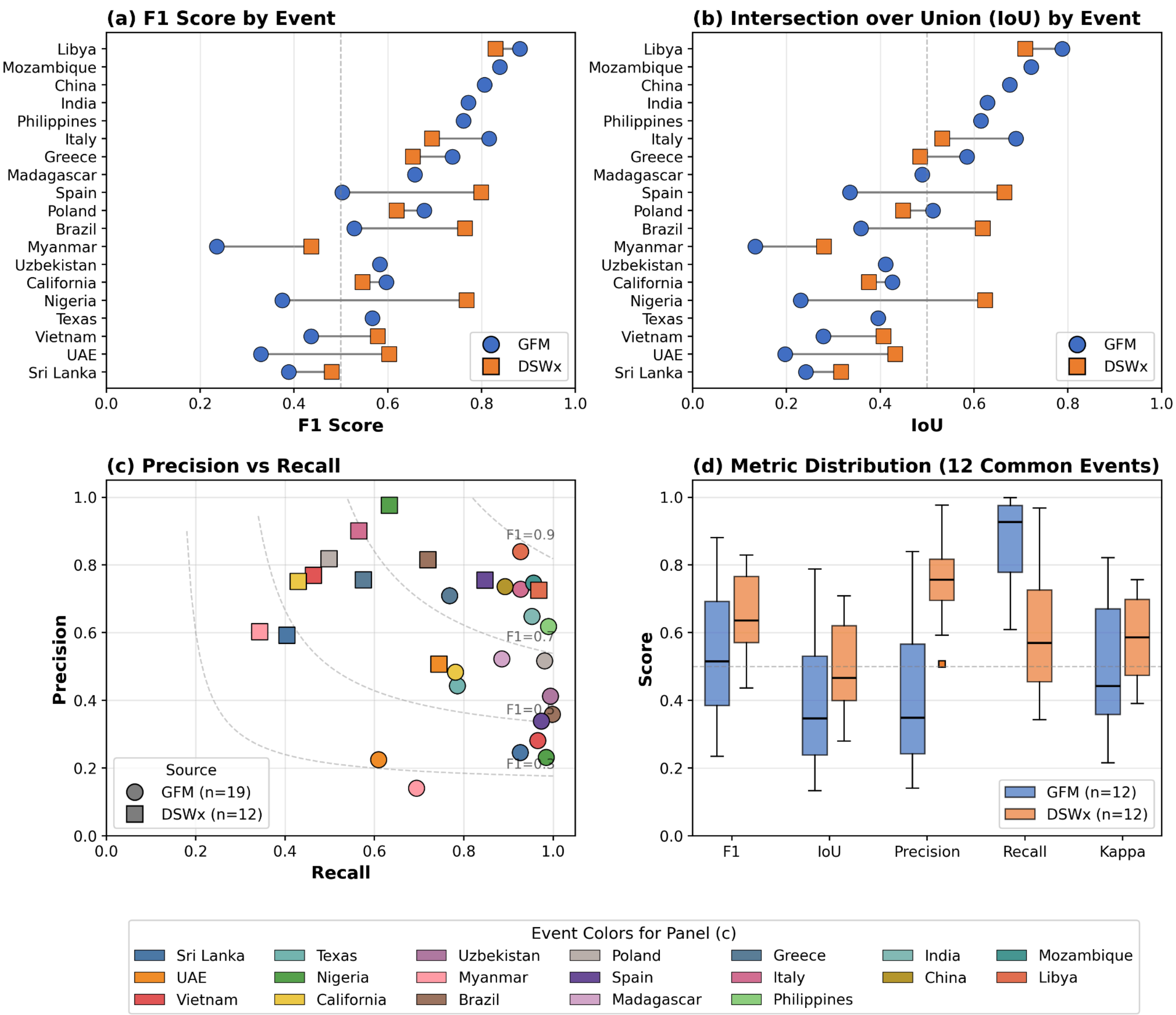

***Figure 3.*** *Per-event validation using Copernicus GFM (n=19) and OPERA DSWx (n=12) references. Panel (a) F1 scores by event, (b) IoU by event, (c) Precision vs. Recall scatter plot with iso-F1 contours at 0.3, 0.5, 0.7, and 0.9; circles=GFM, squares=DSWx, color-coded by event, and (d) Box plots comparing F1, IoU, Precision, Recall, and Kappa distributions across the 12 common events that have both validation sources.*

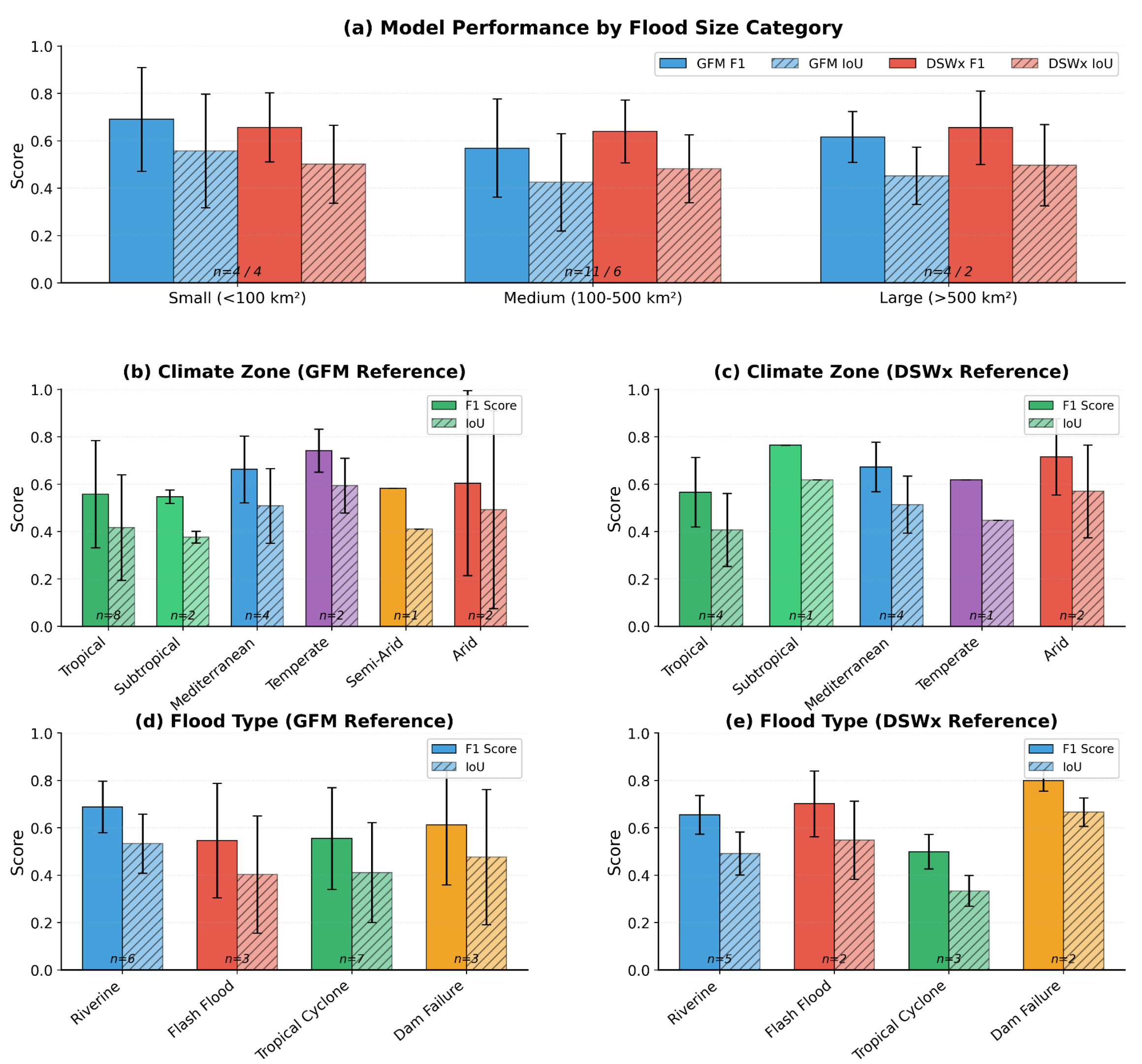

*Figure 4. Model performance stratified by flood size, climate zone, and flood type. (a) F1 and IoU by flood size category (Small <100 km², Medium 100–500 km², Large >500 km²) for both references. (b–c) F1 and IoU by Köppen climate zone for GFM and DSWx references, respectively. (d–e) F1 and IoU by flood mechanism (Riverine, Flash, Cyclone, and Dam Failure) for GFM and DSWx. Error bars show standard deviation; sample sizes are annotated within bars.*

## Stratified Performance by Flood Size, Climate Zone, and Flood Type

We found that detection performance varies systematically with event characteristics. By flood size category, small floods (<100 km², n=4; Fig. 4) achieved the highest mean GFM F1 (0.69 ± 0.22), followed by large floods (>500 km², n=4; F1=0.62 ± 0.11) and medium floods (100 to 500 km², n=11; F1=0.57 ± 0.21; Fig. 4). By Köppen-Geiger climate zone, temperate events (n=2) yielded the highest GFM F1 (0.74 ± 0.09; Fig. 4), followed by Mediterranean (n=4; F1=0.66 ± 0.14) and tropical (n=8; F1=0.56 ± 0.23). By flood type, riverine events (n=6; total area: 7,069.4 km²) achieved the highest GFM F1 (0.69 ± 0.11), followed by dam failure (n=3; 511.4 km²; F1=0.61 ± 0.25), tropical cyclone (n=7;

2,351.9 km²; F1=0.56 ± 0.22), and flash flood (n=3; 645.9 km²; F1=0.55 ± 0.24; Fig. 4). Detailed breakdowns by climate zone, flood type, and DSWx validation are provided in Supplementary Text S2.

The unequal spatial extent across flood types (riverine: 7,069.4 km² across six events vs. flash flood: 645.9 km² across three events) contributes to the observed performance differences, as larger contiguous flood areas provide greater pixel support for detection and yield higher recall. Moreover, the superior riverine performance is consistent with the composition of the Sen1Floods11 training data, which predominantly represents riverine and monsoon-driven flooding. Tropical cyclone events performed worst due to high cloud cover reducing usable imagery, spatially complex inundation mixing coastal storm surge with inland flooding, and vegetation canopy interference. Against DSWx, dam failure events showed the highest F1 (0.80 ± 0.04, n=2), attributable to the rapid, spatially concentrated inundation that produces a clear spectral contrast between pre- and post-breach conditions. Overall, the stratified analysis reveals that riverine floods over open terrain represent the model's strongest detection regime, while tropical cyclone and flash flood events in vegetated or urbanized landscapes represent its weakest. These patterns are consistent with the visual observations in Section 3.2, where spatially coherent flood bodies over cropland and bare terrain produce the highest agreement with reference products.

**Patch-Level Spatial Analysis**

The stratified analysis described above characterizes performance at the event level but does not capture within-event spatial structure. Therefore, we applied connected component analysis to the 19 flood prediction maps to examine sub-event detection patterns (Fig. 5). We identified 79,885 distinct flood patches across 19 events. The patch size distribution is heavily right-skewed, with 70.8% of patches being smaller than 0.01 km² (approximately 11 pixels at 30 m resolution), 23.7% falling between 0.01 and 0.1 km², and only 4.8% exceeding 0.1 km². Patches larger than 10 km² constituted 0.1% of the total (n=83), and 15 patches exceeded 100 km². The predominance of sub-pixel-scale patches indicates that the majority of model-detected flood features consist of isolated pixel clusters. The mean patch size across events was 0.18 km², ranging from 0.03 km² (Myanmar) to 0.75 km² (Libya).

The fragmentation index (patches per km² of flood area) varied from 1.33 (Libya) to 30.08 (Myanmar), with a mean of 10.0. High-performing events (F1 ≥ 0.7, n=7) exhibited a mean fragmentation index of 4.4, while low-performing events (F1 < 0.5, n=5) had a mean of 15.4. This inverse relationship between fragmentation and detection quality is evident in the prediction maps as Libya and Mozambique show coherent, contiguous predictions (Figs. 8, S2), whereas Myanmar and the UAE show dispersed, scattered detections (Figs. S3, 8). Detection recall increased monotonically with patch size for GFM, with a value of 0.69 for very small patches (<0.1 km²), 0.73 for small (0.1 to 1 km²), 0.80 for medium (1 to 10 km²), and 0.86 for large (10 to 100 km²) patches. For DSWx, the same trend held at lower absolute values: 0.25, 0.42, 0.62, and 0.68. The consistent increase indicates that the model reliably captures large contiguous flood areas but produces more omission errors for small, isolated patches.

Overall, we found that events with fragmentation below 5.0 consistently achieved F1 scores above 0.7, whereas events with fragmentation exceeding 15.0 did not exceed 0.5. For operational deployment, a fragmentation threshold could serve as a confidence indicator to flag predictions requiring manual review.

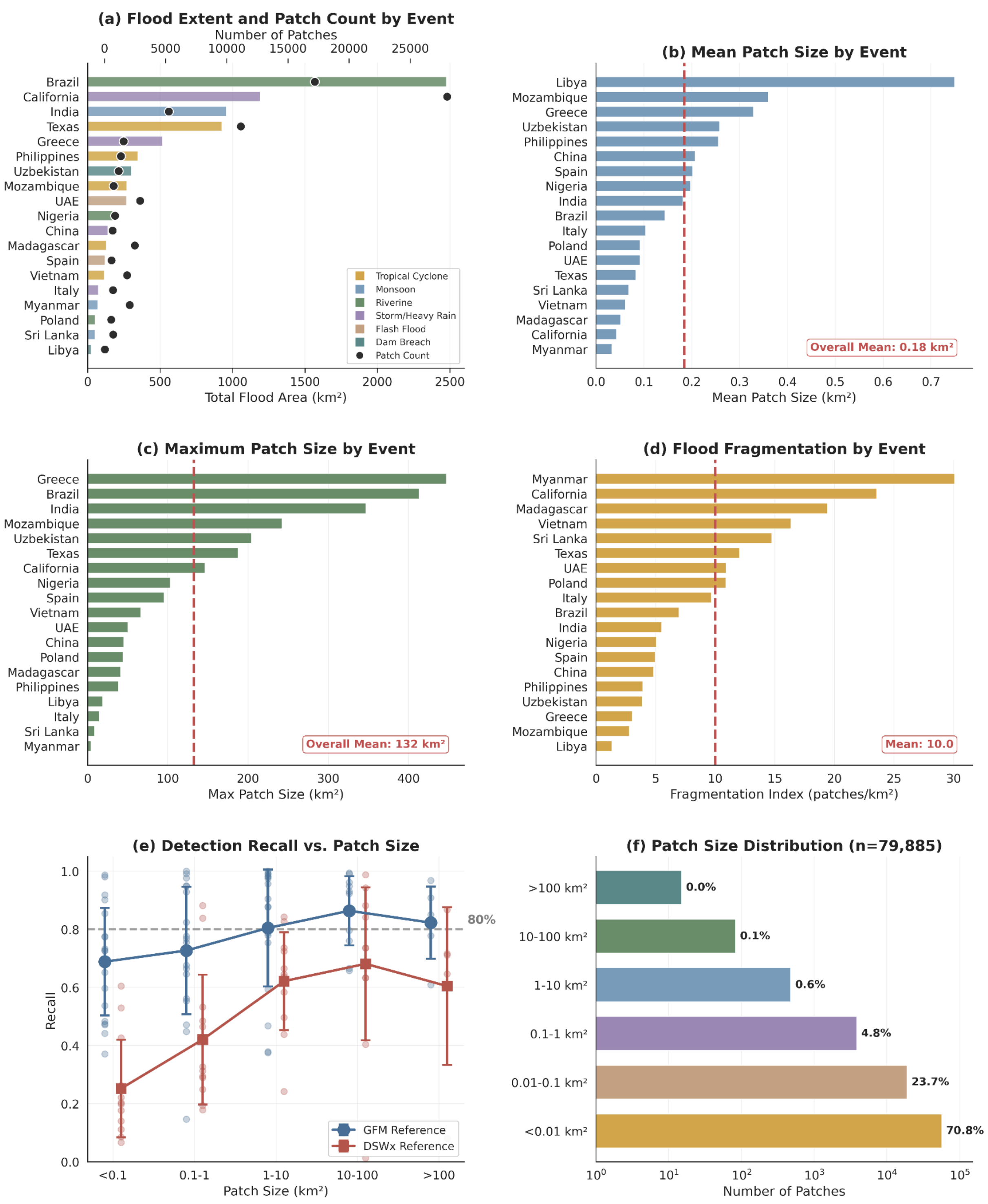

***Figure 5.*** *Patch-level spatial analysis across 19 events. Panel (a) Total flood area (bars, colored by flood type) and patch count (dots) per event. (b) Mean patch size per event. (c) Maximum patch size per event. (d) Fragmentation index (patches/km²) per event. (e) Detection recall by patch size category for GFM and DSWx references; dashed line at 80% recall. (f) Patch size distribution (n=79,885) across six size classes on a log scale.*

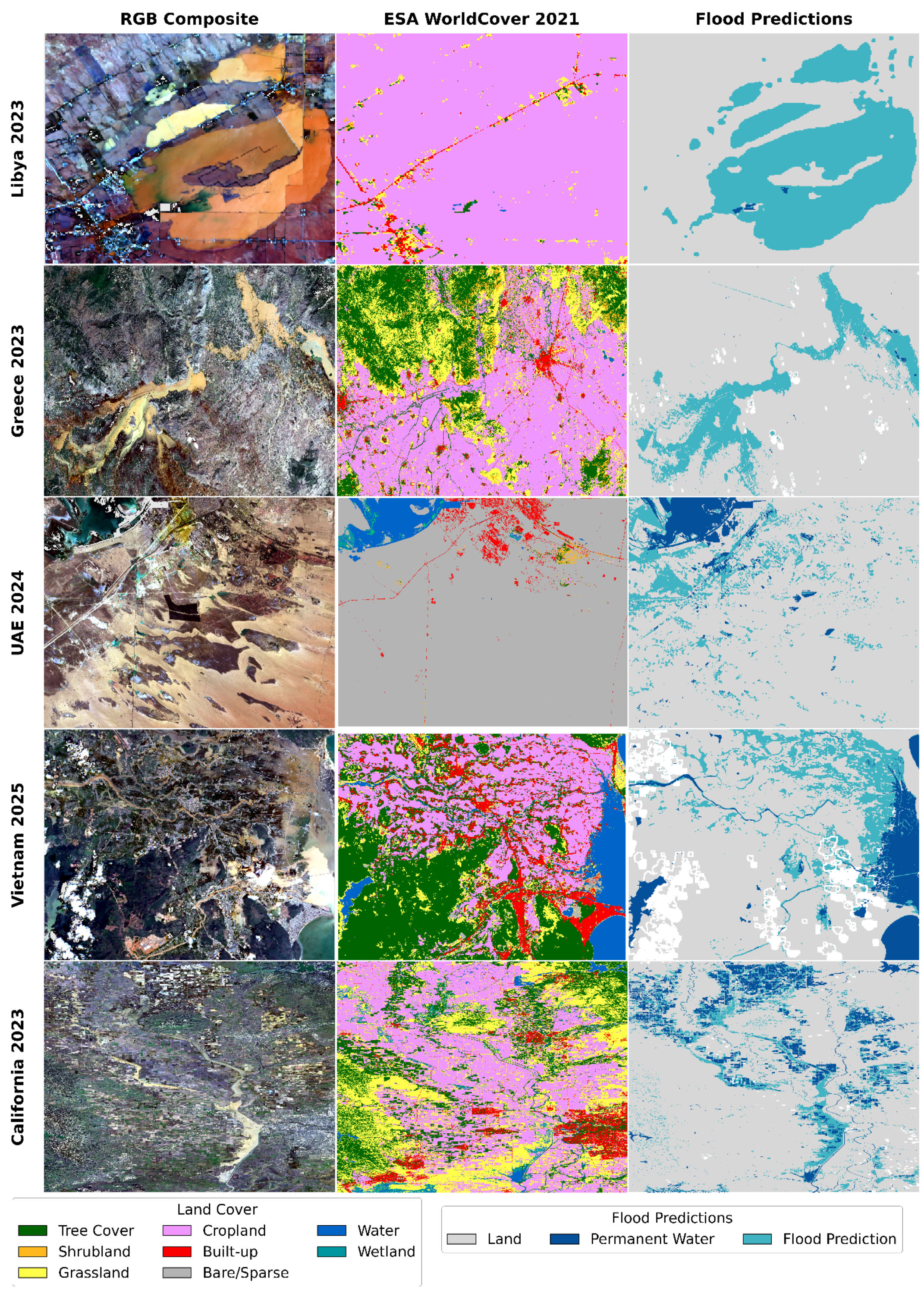

***Figure 6.*** *Input data and flood detection results for five events (Libya 2023, Greece 2023, UAE 2024, Vietnam 2025, and California 2023). Column 1 presents the HLS RGB composite, column 2 presents land cover obtained from the ESA WorldCover 2021 dataset, and column 3 shows Prithvi-EO-2.0 derived flood predictions after postprocessing overlaid with GFM permanent water mask (cyan = flood prediction, dark blue=permanent water, gray=land).*

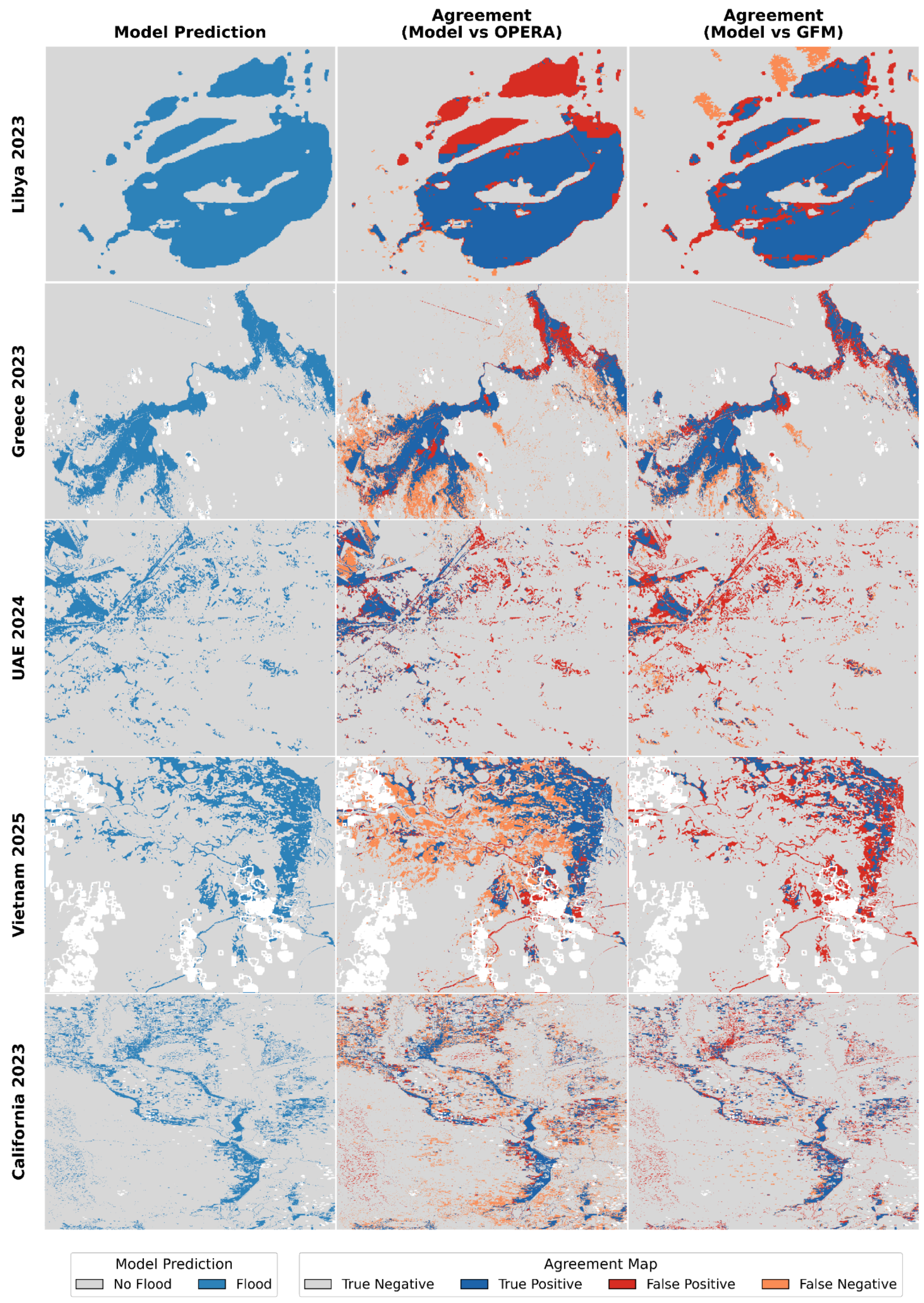

***Figure 7.*** *Pixel-wise agreement maps for the same five events as Figure 6. Column 1 shows Prithvi-EO-2.0 model-derived flood predictions, column 2 presents agreement with OPERA DSWx-HLS, and column 3 depicts agreement with Copernicus GFM. Each color corresponds to four agreement classes: true negative (gray), true positive (blue), false positive (red), and false negative (orange).*

**Visual Comparison and Spatial Validation**

We spatially compared predictions with validation datasets to examine how detection quality relates to local land cover conditions, in addition to per-event metrics that quantify aggregate detection accuracy. We found that flood detection was most spatially coherent over open, low-vegetation land cover classes. For example, in the Libya dam breach event (Fig. 6), the model produced a contiguous flood extent over bare/sparse terrain with minimal fragmentation (32 patches, fragmentation index=1.3; Fig. 5). Events such as Italy (Fig. S1), Greece (Fig. 6), Poland (Fig. S2), India (Fig. S3), and California (Fig. 6) that are cropland-dominated also showed that floodwater along river channels and floodplains is well captured. In contrast, detection quality degrades where tree cover or built-up areas dominate. The Myanmar event (Fig. S3) illustrates the most severe case, driven by heavy cloud cover and cropland interspersed with tree cover. The results depicted a highly fragmented prediction map (fragmentation index = 30.1), with the agreement map (Fig. S5) showing widespread false positives dispersed across the landscape.

We also found that the agreement maps reveal systemic differences between the two references. For Libya (Fig. 7), the OPERA DSWx agreement shows extensive false-positive pixels surrounding the flood core, whereas the GFM agreement shows predominantly true positives in the same area. Conversely, the false-positive rate is higher in GFM than in DSWx for the Nigeria event (Fig. S4). These false positives in GFM occur in cropland and wetland areas, suggesting that the Prithvi-EO-2.0 model detects saturated agricultural surfaces that the SAR-based GFM does not classify as flooded. Furthermore, we found that the flash flood events present distinct spatial challenges. For the UAE (Fig. 7), scattered small flood patches over bare/sparse terrain with built-up areas are confirmed as largely false positives in both references (Fig. 7). These false positives reflect the difficulty of detecting ephemeral flash-flood water in urban environments.

Overall, the visual comparison aligns with the quantitative performance hierarchy, with high-performing events characterized by spatially contiguous flood extents over open terrain (cropland, bare/sparse, grassland), while low-performing events exhibit fragmented predictions in areas dominated by tree cover, built-up surfaces, or heavy cloud interference. False positives tend to concentrate at flood peripheries, in tree cover and built-up zones, and over saturated agricultural surfaces, while false negatives appear primarily in cloud-masked regions and under dense vegetation canopies. The land cover column in Figures 6 and S1-S3 further demonstrates that flood detection is visually most reliable in cropland, bare/sparse, and grassland classes, and least reliable in tree cover and built-up environments, corroborating the quantitative land cover analysis in previous sections.

**Land Cover Effects on Detection Performance**

The visual comparison and patch analysis showed that detection quality varies with local land cover and that spatial coherence is linked to performance. To further quantify this land cover dependence, we stratified the detection metrics by ESA WorldCover class (Fig. 8). Against GFM, cropland yielded the highest IoU (52%) and F1 (69%), followed by wetland (IoU=45%, F1=62%) and grassland (IoU=41%, F1=58%). Built-up areas (IoU=4%, F1=9%) and tree cover (IoU=4%, F1=9%) showed the lowest performance. The strong cropland performance reflects the spectral distinctness of water against bare or low-vegetation agricultural surfaces, conditions well represented in the Sen1Floods11 training data, which include multiple agricultural floodplain events. The near-zero performance on built-up surfaces indicates a fundamental limitation of urban environments, producing spectral mixtures of water, impervious surfaces, and building shadows that the model cannot resolve at 30 m resolution. Similarly, metrics are poor for tree cover, which prevents detection because the canopy obscures underlying inundation from the optical sensor.

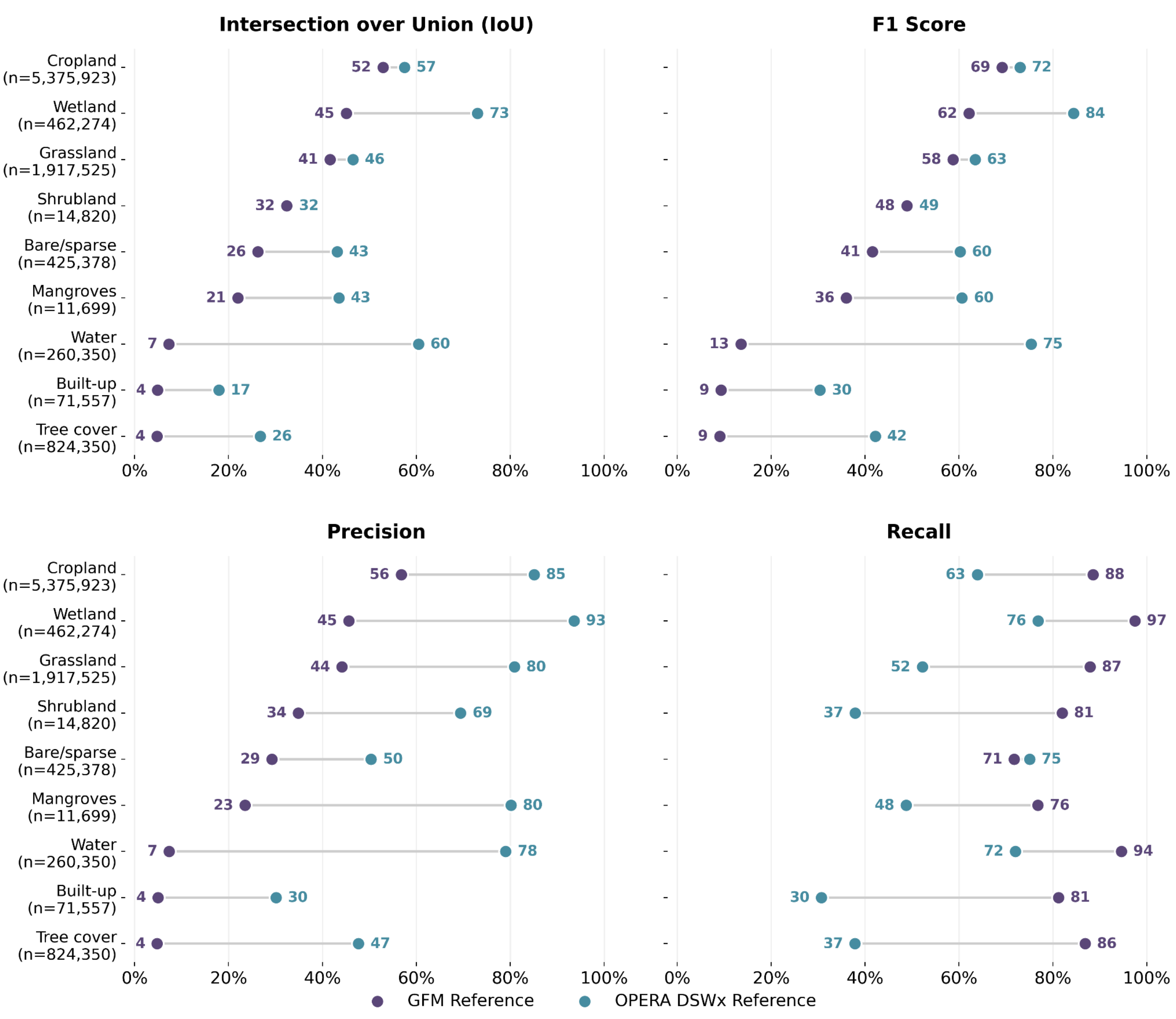

***Figure 8.*** *Model performance by ESA WorldCover land cover class for GFM (purple) and DSWx (teal) references across four metrics: IoU (top left), F1 Score (top right), Precision (bottom left), and Recall (bottom right). Metrics are derived for nine land cover classes: Cropland, Wetland, Grassland, Shrubland, Bare/sparse, Mangroves, Water, Built-up, and Tree cover. Sample sizes per class are indicated in parentheses.*

Against DSWx, the performance hierarchy shifted with wetland achieving the highest IoU (73%) and F1 (84%), followed by permanent water (IoU=60%, F1=75%) and cropland (IoU=57%, F1=72%). The strong wetland performance against DSWx reflects the fact that DSWx maps all surface water, including standing water in wetlands, which is spectrally similar to what the model detects. Figure 9 further stratifies IoU by land cover across six flood types. Cropland IoU ranged from 44% (hurricane and dam breach) to 63% (flash flood). The most striking interaction was grassland, where IoU reached 74% during cyclone events but dropped to 3% during flash floods, suggesting that flood dynamics interact with land cover to determine detection accuracy. This significant contrast might also occur because cyclone-driven flooding in grassland produces extensive shallow inundation with high spectral contrast, whereas flash flooding in grassland is spatially transient and spectrally mixed with sediment. Built-up areas consistently showed near-zero IoU across all flood types (0-12%), confirming that the urban-detection limitation is structural and independent of the flood mechanism.

Overall, we found that cropland, grassland, and bare/sparse terrain represent the model's reliable detection envelope, while tree cover and built-up areas represent hard boundaries where the optical approach cannot be expected to succeed without complementary SAR observations or higher-resolution data. These quantitative findings align with the visual evidence in Figures 6, 7, and S1-S5, where detection coherence maps directly onto the WorldCover classification.

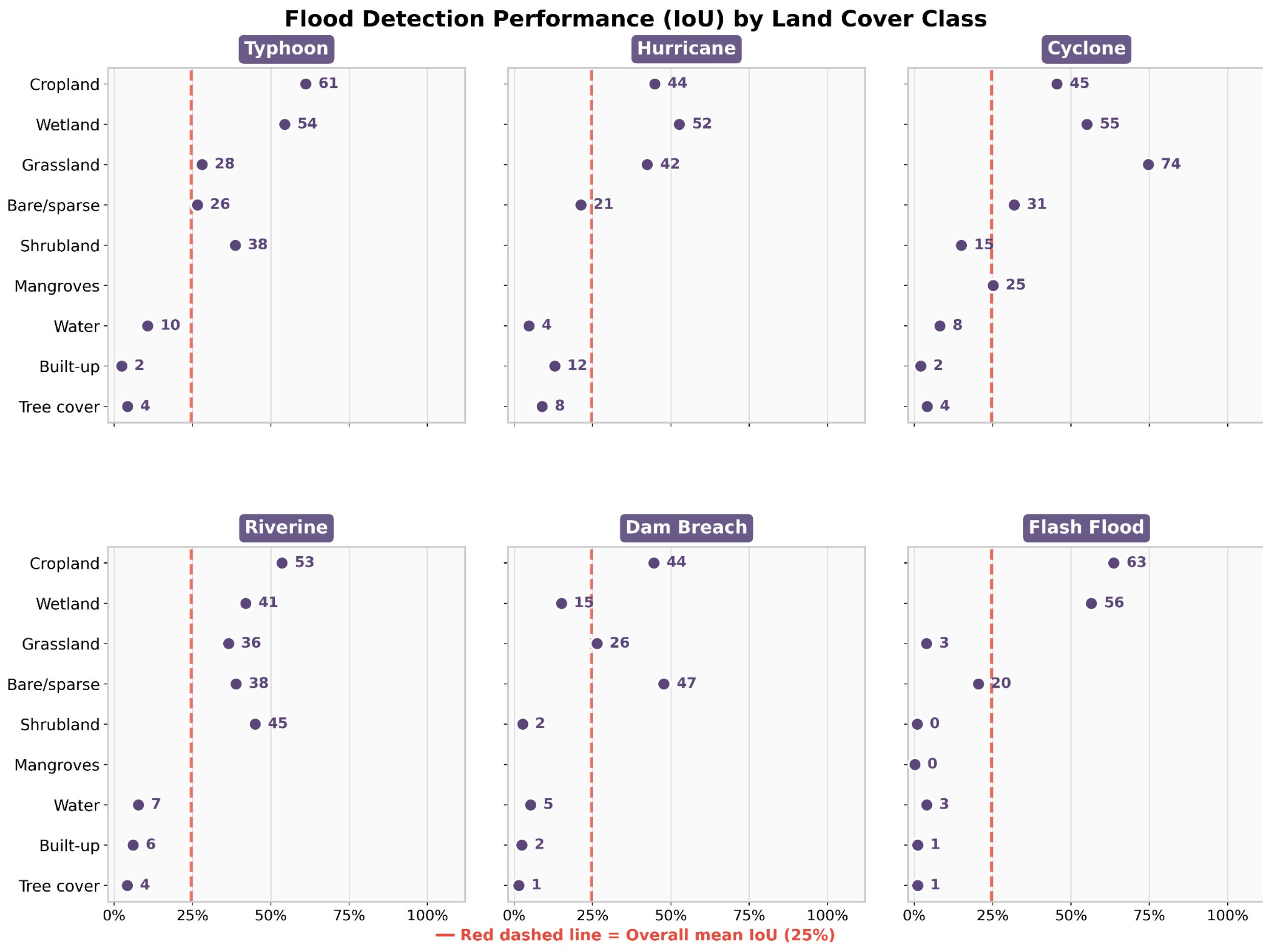

***Figure 9.*** *Flood detection performance assessed using IoU (%) by land cover class stratified across six flood types (Typhoon, Hurricane, Cyclone, Riverine, Dam Breach, Flash Flood) and evaluated against*

*GFM. Each subplot shows IoU for nine land cover classes with values annotated. Vertical red dashed line marks the overall mean IoU (25%).*

**Red-Teaming Results and Failure Mode Taxonomy**

The preceding subsections characterize model performance in terms of detection accuracy. This subsection examines the pipeline itself and documents the failure modes identified through red-teaming during iterative development across 19 events. The evaluation identified 23 distinct failure modes across four pipeline stages, each characterized by root cause, observable symptoms, severity, and validated mitigation (Table S1 and Fig. 10).

Data acquisition and preprocessing accounted for eight failure modes and 42% of pipeline interruptions. Model inference and spatial processing accounted for six failure modes (23%), postprocessing for five failure modes, and validation and reference alignment for four failure modes (12% of interruptions but requiring the most development effort). The complete failure mode taxonomy with root causes, symptoms, and mitigations is provided in Supplementary Table S1.

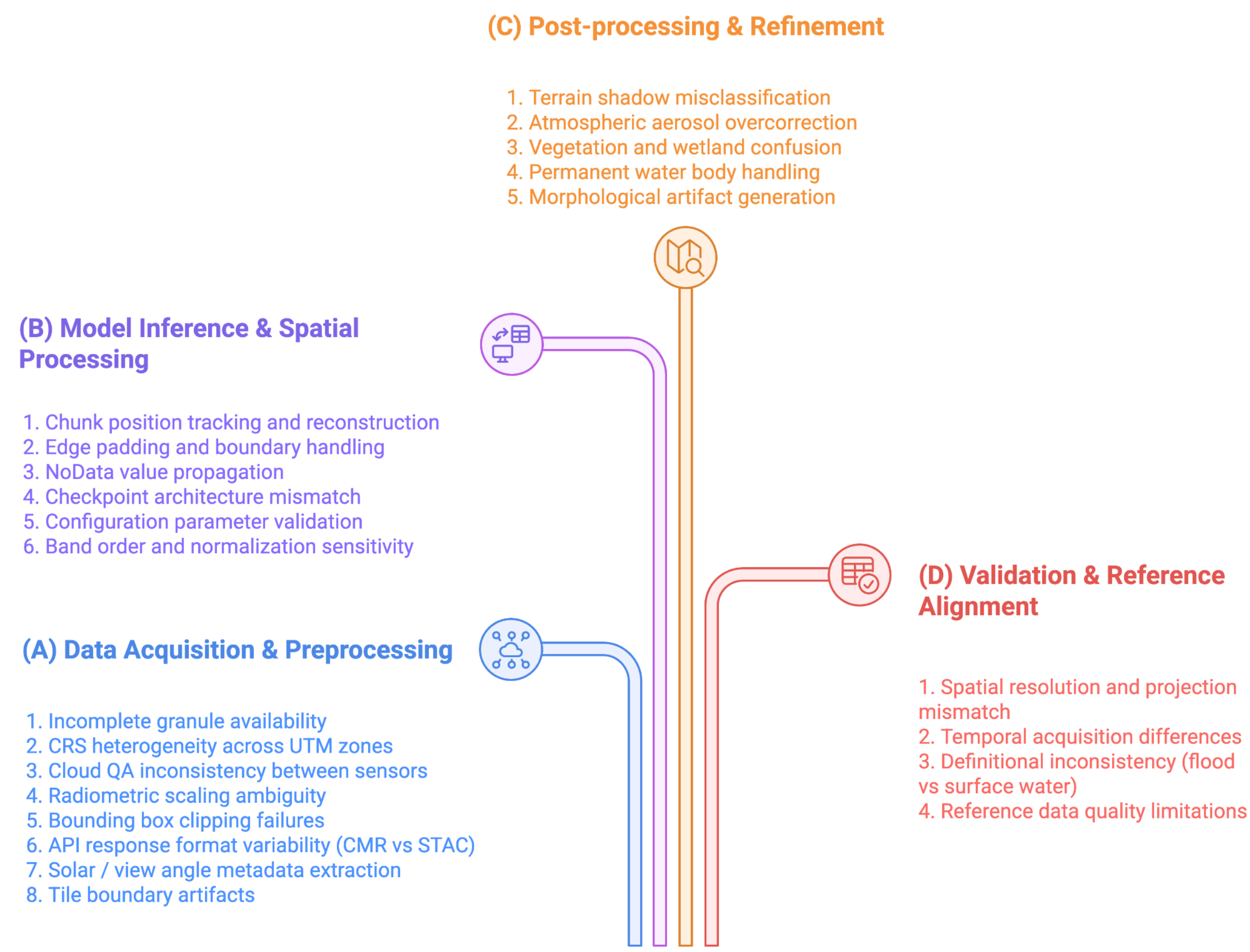

***Figure 10.*** *Taxonomy of 23 failure modes identified during red-team evaluation, organized into four pipeline categories: (a) Data Acquisition & Preprocessing, (b) Model Inference & Spatial Processing, (c) Post-processing & Refinement, and (d) Validation & Reference Alignment.*

The cumulative impact of failure mode mitigation was substantial. Initial validation against OPERA DSWx for the Brazil event produced precision of 4.9% and recall of 4.6%, reflecting the combined effects of spatial misalignment, temporal mismatch, and definitional inconsistency. Systematic resolution of each failure mode progressively improved performance to 82% precision, 72% recall, F1 of 0.77, and IoU of 0.62. This progression demonstrates that the majority of the initial error was attributable to the pipeline implementation rather than the model's capacity. Overall, we found that failure modes exhibit strong interdependencies, such as CRS mismatch propagating through inference to cause validation alignment failures (D1), and NoData handling inconsistencies (B3) interacting with terrain shadow correction (C1) to produce cascading errors. We also observed that validation against existing operational products introduces fundamental limitations when model objectives differ from reference specifications. The initial near-zero precision and recall against OPERA DSWx reflected a definitional mismatch rather than model failure.

Based on operational experience, we recommend five practices for deploying geospatial foundation models, detailed in Supplementary Text S3, including mandatory preprocessing validation, grid-based spatial processing, multi-reference validation frameworks, comprehensive processing logs, and automated regression testing.

**Computational Performance and Operational Deployment**

To support operational use, the pipeline was deployed as a web application (https://explore.fm.odsi.io/), streamlining data download, preprocessing, inference, postprocessing, and visualization for dynamic interaction. The end-to-end pipeline was profiled across all 19 events under two configurations: a sequential research implementation on an NVIDIA V100 GPU workstation and the optimized web deployment (Table S2). In the sequential configuration, mean total processing time was 288.2 seconds (4.8 minutes) per event, ranging from 95.1 seconds (Mozambique, 4 chips) to 1,053.3 seconds (Brazil, 90 chips). In the optimized web deployment, mean processing time reduced to 29.9 seconds per event (range: 19.4 to 54.0 seconds), representing a mean 8.6x reduction. The sub-minute processing time confirms that computational throughput is not the operational bottleneck; the primary latency constraint is HLS data availability, typically two to three days after satellite acquisition. Detailed timing breakdown is provided in Supplementary Text S3 and Table S2.

## Discussion and Conclusions

The mean GFM [DSWx] IoU of 0.46 [0.49] reported across 19 [12] OOD events represents a substantial generalization gap relative to the in-sample Sen1Floods11 benchmark (IoU=0.83 for the water class)[17,18]. This performance or IoU degradation exceeds the approximately 5% mIoU loss reported by Li et al. (2023)[20] when evaluating Prithvi-EO-1.0 on a single OOD event (Bolivia), reflecting the greater geographic, climatic, and mechanistic diversity of the 19-event evaluation set. This OOD degradation pattern is further evident in other foundation model evaluations, where Marsocci et al. (2025)[35] demonstrated, through the PANGAEA benchmark, that GFMs do not consistently outperform supervised baselines such as U-Net across diverse downstream tasks. Similarly, Kaushik et al. (2026)[22] evaluated Prithvi-EO-2.0 on the FloodPlanet benchmark using leave-one-region-out cross-validation with Sentinel-2. They reported a site-averaged mIoU of 0.63, dropping to 0.49 for the Spain hold-out site, where dense urban flooding posed the primary challenge. Our Spain event yielded an F1 of 0.5 and IoU of 0.34 against GFM after postprocessing, encountering similar difficulty attributable to the urban land-cover limitation identified in both studies. Additional details on cross-study comparisons, dual-reference validations, and associated uncertainties are provided in Supplementary Text S4 and S5. These cross-study comparisons confirm that foundation model OOD degradation is a general phenomenon, not specific to Prithvi-EO-2.0, and that its magnitude scales with evaluation diversity.

We found that the model generalizability and transferability varied significantly across the eight climatic regions considered. Events in hot desert and Mediterranean zones yielded the highest detection accuracy (Libya F1=0.88, Italy F1=0.82), while humid subtropical and tropical monsoon zones yielded the weakest (Myanmar F1=0.24, Philippines F1=0.39). Though the model generalized effectively to regions absent from the Sen1Floods11, including Libya, Mozambique, and Italy, the performance degraded in environments where dense vegetation or built-up land cover obscured floodwater. These results indicate that the primary constraint on cross-region transferability is not geographic novelty but the land cover composition and atmospheric conditions at the target site. Events in open terrain with low cloud cover yielded consistent performance regardless of geographic location, whereas events in forested, urban, or cloud-affected regions degraded irrespective of whether similar regions appeared in the training distribution.

The differences in the observed results across events raise the question of whether the observed failures are specific to Prithvi-EO-2.0's architecture and training or represent inherent constraints of optical flood detection at 30 m resolution. We observed that the model achieved an F1 score above 0.8 for events in open terrain (Libya, Mozambique, Italy), indicating that the pretrained representations transfer effectively when the target signal is spectrally accessible. The consistent failure in tree cover and built-up areas, which is also reported for SAR-based models[13] and for alternative GeoFMs on the same datasets[22], suggests that most of these detection boundaries are structural constraints of the observation geometry and spatial resolution rather than deficiencies in the Prithvi encoder or its pretraining strategy.

Nevertheless, certain aspects of the failure pattern are attributable to model and training data design rather than just sensor physics. The Sen1Floods11 dataset (446 labeled chips from 11 events) used for fine-tuning predominantly represents riverine floods in open terrain. It lacks or underrepresents events related to urban flooding, flash floods in mountainous terrain, and tropical cyclone-induced inundation under heavy cloud cover. The model's weak performance on these event types, therefore, reflects both optical detection limits and compositional bias in the fine-tuning dataset. Additionally, Prithvi-EO-2.0 ingests six HLS bands without explicit water index features such as Normalized Difference Water Index (NDWI) or Modified NDWI (MNDWI). While the model can, in principle, learn equivalent spectral combinations through self-supervised pretraining, the absence of these indices as explicit input channels may reduce sensitivity to subtle water-land boundaries in spectrally mixed pixels. As Prithvi-EO-2.0 was already pretrained on an archive of 4.2 million HLS chips that covers broad global coverage across land cover types, climate zones, and seasons, additional pretraining is therefore unlikely to address the detection gaps identified here. However, since flood detection capability is acquired entirely during downstream fine-tuning, increasing the scale of more diverse labeled flood data, particularly from underrepresented event types and land cover classes, might improve accuracy.

The red-teaming methodology identified 23 failure modes across four pipeline stages, with preprocessing accounting for 42% of all interruptions. Initial validation yielded a precision of 4.9% and a recall of 4.6%, improving to 82% and 72%, respectively, after systematic failure mode resolution. This progression demonstrates that pipeline implementation, not model capacity, dominated initial error. Li et al. (2026)[27] reached an analogous conclusion in the domain of flood forecasting. When Google's AI flood prediction system[36] was re-evaluated using observation-based rather than model-based thresholds, the median F1 for 5-year floods dropped from approximately 0.42 to below 0.20, and false negative rates for extreme events exceeded 90%. In both cases, evaluation methodology and pipeline configuration contributed more to reported accuracy than model architecture itself. From the model capability and deployment standpoint, we recommend the following to the practitioners evaluating Prithvi-EO-2.0 for operational flood mapping. The model can be deployed with confidence for riverine and dam failure events over cropland, grassland, wetland, and bare terrain in arid, semi-arid, and temperate climate zones, where it consistently achieves F1 above 0.7 without event-specific training or manual intervention. Processing time of under one minute per event through the deployed web portal enables rapid situational awareness during disaster response. However, the model should not be used as a sole detection source for flash floods, tropical cyclone events, or any event where tree cover or built-up land cover dominates the affected area, as detection accuracy in these conditions approaches zero. For these events, SAR-based products such as Copernicus GFM or manual interpretation of higher-resolution imagery remain necessary. The fragmentation index provides an automated quality indicator: events producing a fragmentation index below 5.0 consistently yield reliable maps (F1 > 0.7), while higher values signal that predictions should be verified against independent data sources.

Several limitations constrain the interpretation of this study, including the use of operational reference products rather than independent ground truth, evaluation of a single foundation model, and reliance on a single fine-tuning configuration. Detailed limitations, future design recommendations for foundation models, and future research directions are discussed in Supplementary Text S6. Overall, this study provides the first systematic evaluation of a geospatial foundation model deployed operationally for flood mapping across 19 OOD events globally. The red-teaming methodology identified 23 failure modes, with data preprocessing and pipeline alignment contributing more to initial error than model capacity. The model achieved a mean GFM F1 of 0.61 and IoU of 0.46 across all events, with operationally useful detection (F1 > 0.7) for 10 of 19 events without task-specific fine-tuning beyond the Sen1Floods11 adaptation. Detection performance exhibited strong land cover dependence, with cropland, grassland, and bare terrain forming the reliable detection envelope, while tree cover and built-up areas represented the structural limits of the optical approach. Dual-reference validation against SAR-based (GFM) and optical-based (DSWx-HLS) products revealed that apparent model error substantially reflects definitional inconsistency between flood water and surface water rather than detection failure. These findings indicate that AI-based flood model evaluations confined to benchmark settings underestimate the engineering complexity of operational deployment, a pattern that extends beyond foundation models for flood mapping to AI-based flood forecasting systems more broadly. The failure mode taxonomy, dual-reference validation framework, and land-cover-stratified performance characterization presented here are applicable to future GeoFM assessments for flood mapping and other Earth observation tasks. Multimodal fusion of optical and SAR inputs, multi-dataset fine-tuning, and higher-resolution sensors represent pathways toward resolving the detection gaps identified in this study. Overall, the results establish that Prithvi-EO-2.0 provides operationally reliable flood detection for riverine events over open terrain but cannot serve as a sole detection source for flash floods or tropical cyclone events in vegetated or urban environments. These detection limits are predominantly structural constraints of 30 m optical imaging rather than model-specific deficiencies, indicating that progress requires multi-sensor integration with SAR and diversification of labeled training data across underrepresented flood regimes rather than architectural changes to the foundation model.

## Data and Methods

This study quantifies the operational reliability of a GeoFM for flood inundation mapping by testing the entire processing pipeline across 19 OOD flood events. The workflow encompasses: (a) selection of spatiotemporally diverse flood events, (b) HLS data acquisition and preprocessing, (c) applying Prithvi-EO-2.0 GeoFM fine-tuned for flood extent mapping, (d) physics-based postprocessing, (e) pixelwise validation against two independent reference products and (f) stratified performance analysis. An iterative red-teaming methodology was applied throughout deployment to identify and mitigate failure modes at each pipeline stage (Fig. 2).

### Flood Event Selection and Study Area

We selected 19 flood events for OOD evaluation of the Prithvi-EO-2.0 GeoFM with no geographic or temporal overlap with the 11 events [Bolivia, Ghana, India, Mekong, Nebraska (USA), Nigeria, Pakistan, Paraguay, Somalia, Spain, and Sri Lanka between 2016 and 2019] in the Sen1Floods11 training dataset[19]. We have also selected the events to maximize diversity along four dimensions: geographic distribution across six continents, temporal range covering most HLS availability [2017 to 2025], representation of six flood mechanisms (riverine, flash, dam breach typhoon-, hurricane-, and cyclone-induced), and variation in flood spatial extent from sub-100 km² to >10,000 km². Additionally, we have selected events that resulted in documented socioeconomic losses, including population displacement, infrastructure damage, and economic losses. This ensures that the evaluation considers operationally relevant flood events rather than transient inundation events. Event locations, metadata, and spatial characteristics are summarized in Tables 1 and 2 and Fig. 1.

### Data Acquisition

We obtained the Harmonized Landsat and Sentinel-2 imagery at 30 m resolution for all 19 events using the Common Metadata Repository (CMR) API for granules before 2023 and the SpatioTemporal Asset Catalog (STAC) API for post-2023 acquisitions[37]. We acquired six spectral bands [Blue, Green, Red, Near-Infrared, Shortwave Infrared 1, and Shortwave Infrared 2] required for Prithvi-EO-2.0 model inferencing, along with the Fmask QA band for cloud and shadow masking (Section 2.5), solar zenith angle (SZA) and solar azimuth angle (SAA) bands for terrain shadow correction (Section 2.6).

Model predictions were validated against two independent operational reference datasets. We obtained the Copernicus Global Flood Monitoring (GFM) dataset that comprises near-real-time flood extent maps derived from Sentinel-1 C-band Synthetic Aperture Radar (SAR) imagery at 10 m resolution[38]. For validation, GFM flood extent maps were reprojected and resampled to the 30 m HLS prediction grid using nearest-neighbor interpolation to preserve binary classification values. GFM integrates three independently developed flood-mapping algorithms (HASARD, TU Wien, and DLR) into an ensemble product, where pixel-wise flood classifications are determined by majority voting across algorithm outputs. The product distinguishes between observed flood extent (transient inundation exceeding the reference water mask) and a reference water mask representing permanent water bodies. GFM data were available for all 19 events in this study, accessed via the STAC API at the Earth Observation Data Centre (EODC). The GFM permanent water mask was additionally used as an auxiliary input for the postprocessing pipeline (Section 2.6). Since GFM incorporates SAR imagery that operates independently of cloud cover and illumination conditions, it provides a complementary sensor modality relative to the optical HLS inputs used by Prithvi-EO-2.0.

The second validation dataset was obtained from the Observational Products for End-Users from Remote Sensing Analysis (OPERA) Dynamic Surface Water Extent (DSWx) from the Harmonized Landsat and Sentinel-2 (DSWx-HLS) product that provides maps of surface water extent at 30 m resolution derived from HLS optical imagery[39]. A critical distinction from GFM is that DSWx-HLS maps total surface water extent rather than flood water specifically, i.e., it does not differentiate between permanent water bodies and transient flood inundation. DSWx-HLS data were available for 12 of the 19 events, with availability constrained by the product's operational start date and cloud-free observation requirements. The rationale for employing two reference products stems from the absence of a single authoritative ground truth for flood extent at the spatial and temporal scales considered in this study. Each product introduces distinct biases, and cross-comparison enables identification of systematic discrepancies attributable to definitional differences rather than model error.

### Data Preprocessing Pipeline

The preprocessing pipeline converts raw HLS granules into model-ready input tensors through four sequential operations (Fig. 2, Panel B) informed by preprocessing failure modes identified during iterative red-teaming (Section 2.8). These failures caused either silent degradation of predictions or complete pipeline interruption. First, cloud and cloud shadow masking is applied using the HLS Fmask QA bands[37] with affected pixels excluded from subsequent processing and inference. Second, all granules are reprojected to a common coordinate reference system and merged into a spatially continuous mosaic, with each spectral band stored separately to facilitate downstream band-specific postprocessing.

Third, radiometric normalization applies per-band z-score standardization using the mean and standard deviation scalars computed from the Prithvi-EO-2.0 pretraining dataset. Scene-specific normalization was avoided because it degraded model performance. Fourth, imagery is tiled into 512 × 512-pixel chips with explicit, grid-based position tracking. Edge chips are zero-padded and cropped after inference to ensure seamless reconstruction, avoiding cumulative coordinate errors observed with geospatial transform-based methods. Additional preprocessing includes validating band completeness, verifying radiometric range, and standardizing NoData encoding to a consistent fill value (e.g., -9999) across sensors and excluding NoData pixels from inference to prevent artifact propagation. These checks reject non-conforming inputs before inference, preventing silent failures from propagating through the pipeline.

### Prithvi-EO-2.0 Foundation Model

We selected Prithvi-EO-2.0 for flood mapping among the other recently developed GFMs for three reasons. First, Prithvi-EO-2.0 is one of the largest open-source GeoFMs pretrained on HLS data, with approximately 600 million parameters in a temporal Vision Transformer (ViT) architecture. It also provides direct compatibility with the HLS data pipeline without requiring sensor-specific adaptation[17,18]. Second, the model was developed in-house by the authors at NASA Marshall Space Flight Center and IBM Research. Moreover, the model has been publicly released with pretrained weights and a fine-tuned flood detection checkpoint, making it the most accessible GFM for operational deployment within NASA's Earth science data ecosystem. Third, prior evaluations demonstrated that Prithvi outperforms U-Net on OOD flood data[20], and its architecture has been adopted in subsequent studies[21], establishing it as the benchmark GFM for flood mapping applications.

The model was pretrained using a Masked Autoencoder (MAE) strategy[40] on over 4.2 million HLS image chips sampled globally from the 2013-2024 archive period[18]. This vast dataset enables learning general-purpose spatiotemporal representations from six spectral bands without labeled supervision. The pretrained encoder was subsequently fine-tuned for flood inundation mapping on the Sen1Floods11 dataset[19], which comprises 446 labeled 512 x 512 image chips across 11 flood events. The fine-tuned architecture pairs the ViT encoder with a U-Net decoder that receives multi-scale feature representations from four intermediate encoder layers and produces pixelwise binary flood probability maps. The ViT encoder provides pretrained spatiotemporal representations through masked autoencoding, while the U-Net decoder is trained specifically for flood segmentation during fine-tuning on Sen1Floods11. The 600M model with temporal and location embeddings achieved an IoU of 83.1 for the water class and a micro-averaged F1 of 97.7 during testing[18]. Tulbure et al. (2025)[25] reported that the 600M version yields higher recall but lower precision than the 300M-TL variant, consistent with the high-recall, low-precision pattern observed in our evaluation. It is important to note that no additional fine-tuning or domain adaptation was performed for this study. This design choice ensures that the evaluation reflects the operational scenario in which a researcher or practitioner deploys the model as distributed, without access to labeled data for the target events. A threshold of 0.5 converts probability outputs to binary flood/non-flood classification (Fig. 2, Panel C).

**Physics-Based Postprocessing**

Raw model predictions were refined using a four-step, physics-based postprocessing pipeline to reduce systematic false positives arising from surface features with optical signatures similar to those of water before validation (Fig. 2, Panel D). First, we applied terrain shadow correction to remove flood predictions over topographically shaded surfaces. We computed the Solar Local Incidence Angle (SLIA) using scene-specific solar zenith and azimuth angles from HLS metadata. We also obtained slope and aspect derived from the Copernicus GLO-30 DEM. We then masked pixels predicted as flood with SLIA greater than 85° and slope exceeding 15°, reclassifying them as non-flood. Second, we applied atmospheric aerosol filtering to address false water detections caused by anomalous surface reflectance under high aerosol loading (haze, smoke, dust). We identified aerosol conditions from the Fmask QA band, and pixels classified as flood under high aerosol levels are removed based on their NIR-to-Green reflectance ratio.

Third, we employed vegetation filtering to remove false flood detections over dense vegetation, which can exhibit low reflectance in visible bands under certain illumination or moisture conditions [41]. We computed the Normalized Difference Vegetation Index (NDVI) from the Red and NIR bands, and pixels predicted as flood with NDVI > 0.7 are reclassified as non-flood. In general, such high NDVI values indicate dense photosynthetically active vegetation, inconsistent with surface water [42,43]. Fourth, we applied permanent water body masking to remove detections that do not constitute flood inundation. We used the permanent water mask from the GFM dataset[44] to reclassify from flood to non-flood. The quantitative contribution of each correction step to the total false positive reduction is reported in Section 3.

**Stratified Analysis and Accuracy Metrics**

Pipeline performance was evaluated at both the scene level (per-event aggregate metrics) and the patch level (individual connected flood components). Scene-level validation computes pixelwise confusion matrices for each event against each available reference product and derives standard accuracy metrics, including overall accuracy (OA), precision, recall, F1-score, intersection over union (IoU), and frequency bias. Precision and recall are emphasized for operational relevance, with recall prioritized in disaster response contexts where missed flooding carries greater consequences than false alarms.

We also stratified scene-level metrics by four environmental variables to characterize context-dependent performance, namely, flood spatial extent (area-based size categories), Köppen-Geiger climate zone, flood mechanism (riverine, flash, dam breach typhoon-, hurricane-, and cyclone-induced), and land cover class (derived from 10 m ESA WorldCover 2021)[45]. This stratification addresses a key gap in prior GFM evaluations, which typically report aggregate metrics without disaggregating by the environmental factors known to influence flood behavior and detection difficulty.

We further implemented patch-level analysis to decompose each event's flood prediction into individual connected components (flood patches) and evaluate detection recall as a function of patch size and fragmentation. We identified flood patches in both the prediction and the reference using connected-component labeling and computed the spatial overlap between the predicted and reference patches. This analysis quantifies the minimum reliably detectable flood patch size, a critical parameter for operational interpretation of flood maps in disaster response settings.

All analysis was implemented in Python using PyTorch for model inference (via the PyTorch Geometric framework), rasterio and rioxarray for geospatial operations, scipy for terrain analysis, and data discovery and access using the Common Metadata Repository (CMR) API for granules before 2023 and the SpatioTemporal Asset Catalog (STAC) API via the pystac_client library for post-2023 acquisitions. Model inference utilized the publicly available Prithvi-EO-V2-600-Sen1Floods11 checkpoint. All experiments were conducted on a workstation equipped with an NVIDIA GPU. Processing time per event is reported in Section 3.

**Red-Teaming Methodology**

The preprocessing and postprocessing, along with validation strategies described above, represent the final, refined end-to-end pipeline. These components within each stage were not predetermined but emerged through iterative testing that exposed failure modes at each pipeline stage. We applied red teaming, a structured adversarial testing methodology originating in military and cybersecurity applications and recently adopted for AI system evaluation[29], to systematically identify the taxonomy of failure modes not captured during controlled benchmark testing. The approach involved iterative deployment of the complete pipeline across diverse flood events, with deliberate variation in geographic region, sensor platform (Landsat vs. Sentinel-2), atmospheric conditions, land cover, and event temporal proximity to the training data.

Each identified failure mode was documented using four attributes: root cause, observable symptoms, impact severity, and mitigation approach. Failure mode taxonomies were further categorized by pipeline stage, including data acquisition and preprocessing, model inference and spatial processing, postprocessing and refinement, and validation and reference alignment (Fig. 10). This categorization enables attribution of failures to specific pipeline components and informs targeted quality control development. The four stages form a sequential processing chain in which failures propagate downstream; for example, CRS inconsistency in data acquisition (Stage A) can cascade through inference (Stage B) to produce spatial misalignment in validation (Stage D). We also want to note that red-teaming was conducted iteratively rather than as a single-pass exercise. After implementing each mitigation, we reprocessed all previously evaluated events to verify that corrections did not introduce new failures or degrade performance. This regression-testing strategy, adapted from software engineering practices, ensured the pipeline's cumulative stability.

***Table 1*** *Summary of 19 flood events used for model evaluation. Events marked with an asterisk (*) have two independent validation sources (Copernicus GFM and OPERA DSWx-HLS); remaining events are validated against GFM only. (Note: DANA - Depresión Aislada en Niveles Altos)*

| ID | Event | Location | Date | Flood Type | Cause | Deaths | Damage |
|---|---|---|---|---|---|---|---|
| 1 | Brazil* | Porto Alegre, Rio Grande do Sul | 2024-05-06 | Riverine | Extreme rainfall | 183 | $15B+ |
| 2 | California* | Sacramento Valley, Davis | 2023-01-22 | Riverine | Atmospheric Rivers | 22 | $3B+ |
| 3 | Greece* | Larissa, Thessaly Plain | 2023-09-10 | Riverine | Storm Daniel | 17 | S4.4B+ |
| 4 | UAE* | Dubai, Abu Dhabi | 2024-04-19 | Flash | Extreme rainfall | 4 | $3B+ |
| 5 | Myanmar* | Magway Region | 2024-09-20 | Typhoon | Typhoon Yagi | 113+ | - |
| 6 | Poland* | Wroclaw, Lower Silesia | 2024-09-18 | Riverine | Storm Boris | 10 | $4.62B |
| 7 | Italy* | Ravenna, Emilia-Romagna | 2023-05-23 | Riverine | Extreme rainfall | 17 | $9.35B |
| 8 | Nigeria* | Maiduguri, Borno State | 2024-09-12 | Dam Breach | Alau Dam collapse | 30 | - |
| 9 | Sri Lanka* | Gampaha District | 2025-11-30 | Cyclone | Tropical cyclone | - | - |
| 10 | Vietnam* | Central Vietnam | 2025-11-07 | Cyclone | Tropical weather | - | - |
| 11 | Spain* | Valencia | 2024-10-30 | Flash | DANA | 228 | $14.3B+ |
| 12 | Libya* | Marj, Cyrenaica | 2023-09-15 | Dam Collapse | Storm Daniel | 5,000+ | - |
| 13 | Texas | Houston Metro Area | 2017-08-30 | Hurricane | Hurricane Harvey | 107 | $125B |
| 14 | India | Patna, Bihar | 2019-09-23 | Riverine | Monsoon flooding | 130 | - |
| 15 | Philippines | Amulung, Cagayan | 2020-11-13 | Typhoon | Super Typhoon Vamco | 67 | $270M |
| 16 | Uzbekistan | Sardoba Dam, Syrdarya | 2020-05-04 | Dam Breach | Dam failure | 6 | $1B+ |
| 17 | Madagascar | Antananarivo | 2022-01-30 | Cyclone | Tropical Storm Ana | 55 | $75M+ |
| 18 | China | Xinxiang, Henan Province | 2021-07-26 | Flash | Extreme rainfall | 398 | $18.6B |
| 19 | Mozambique | Beira, Sofala Province | 2019-03-22 | Cyclone | Cyclone Idai | 602 | $3.3B |

***Table 2*** *Spatial characteristics of the 19 flood events. Bounding box coordinates are in decimal degrees (lon_min, lat_min, lon_max, lat_max). Chips denote the number of 512×512 pixel tiles. Flood (%) represents the proportion of valid areas classified as flood. Cloud and Shadow percentages indicate data loss from Fmask-based masking. Events marked with an asterisk (*) have two independent validation sources.*

| ID | Event | Bounding Box | Chips | Total Area (km²) | Flood Area (km²) | Flood (%) | Cloud (%) | Shadow (%) |
|---|---|---|---|---|---|---|---|---|
| 1 | Brazil* | -53.42, -30.41, -51.19, -29.54 | 90 | 23744 | 3075.86 | 15.32 | 1.3 | 0.96 |
| 2 | California* | -122.17, 38.17, -121.38, 39.60 | 45 | 13906 | 1913.13 | 18.27 | 0.31 | 0.71 |
| 3 | Greece* | 21.75, 39.38, 22.68, 39.81 | 18 | 3651.89 | 489.8 | 13.63 | 0.41 | 1.16 |
| 4 | UAE* | 54.23, 23.95, 54.84, 24.32 | 15 | 2814 | 371.12 | 14.61 | 0 | 0 |
| 5 | Myanmar* | 94.71, 20.17, 94.94, 20.84 | 10 | 1815.8 | 199.79 | 18.56 | 32.28 | 8.17 |
| 6 | Poland* | 18.14, 49.93, 18.35, 50.33 | 6 | 1417.62 | 95.61 | 7.24 | 2.92 | 3.9 |
| 7 | Italy* | 11.58, 44.44, 12.10, 44.61 | 8 | 1094 | 76.5 | 10.9 | 0.05 | 0.57 |
| 8 | Nigeria* | 13.12, 11.74, 13.38, 12.06 | 6 | 984.97 | 170.62 | 20.33 | 8.73 | 5.85 |
| 9 | Sri Lanka* | 79.88, 6.86, 80.10, 7.17 | 6 | 843.69 | 49.7 | 7.17 | 5.36 | 12.43 |
| 10 | Vietnam* | 108.97, 13.72, 109.25, 13.96 | 6 | 813.49 | 181.71 | 25.88 | 6.5 | 7.2 |
| 11 | Spain* | -0.45, 39.15, -0.32, 39.47 | 3 | 528 | 141.79 | 40.99 | 4.63 | 6.48 |
| 12 | Libya* | 20.87, 32.49, 20.96, 32.57 | 1 | 80 | 23.32 | 35.18 | 0 | 0 |
| 13 | Texas | -96.86, 28.88, -95.42, 29.67 | 60 | 12489.93 | 1039.15 | 9.3 | 6.49 | 4.03 |
| 14 | India | 84.18, 25.59, 85.23, 26.05 | 32 | 5853 | 1418.49 | 33.23 | 7.12 | 9.92 |
| 15 | Philippines | 121.61, 17.54, 121.78, 18.32 | 12 | 1647 | 390.27 | 57.52 | 47.47 | 8.83 |
| 16 | Uzbekistan | 68.21, 40.28, 68.67, 40.67 | 9 | 1625.5 | 317.45 | 20.1 | 0.7 | 2.15 |
| 17 | Madagascar | 47.27, -19.06, 47.67, -18.72 | 9 | 1571.5 | 135.3 | 12.48 | 15.7 | 15.32 |
| 18 | China | 114.17, 35.44, 114.50, 35.75 | 9 | 1257 | 133.01 | 17.59 | 8.79 | 11.75 |
| 19 | Mozambique | 34.53, -20.22, 34.78, -19.99 | 4 | 686.07 | 356.03 | 70.64 | 15.25 | 11.29 |

***Table 3*** *Per-event validation metrics for the Prithvi-EO-2.0 flood model against Copernicus GFM (n=19) and OPERA DSWx-HLS (n=12) references. Events are grouped by availability of both references (top, sorted by GFM IoU descending), followed by GFM-only events (bottom, same sorting). Dash (–) indicates that the DSWx reference is not available.*

| **Event** | **GFM Accuracy** | **DSWx Accuracy** | **GFM F1** | **DSWx F1** | **GFM IoU** | **DSWx IoU** | **GFM Precision** | **DSWx Precision** | **GFM Recall** | **DSWx Recall** | **GFM Bias** | **DSWx Bias** |
|---|---|---|---|---|---|---|---|---|---|---|---|---|
| Libya | 92.05 | 89.55 | 0.88 | 0.83 | 0.79 | 0.71 | 0.84 | 0.73 | 0.93 | 0.97 | 1.10 | 1.33 |
| Italy | 96.73 | 92.25 | 0.82 | 0.69 | 0.69 | 0.53 | 0.73 | 0.90 | 0.93 | 0.57 | 1.27 | 0.63 |
| Greece | 93.16 | 89.12 | 0.74 | 0.65 | 0.59 | 0.49 | 0.71 | 0.76 | 0.77 | 0.58 | 1.08 | 0.76 |
| Poland | 96.42 | 92.87 | 0.68 | 0.62 | 0.51 | 0.45 | 0.52 | 0.82 | 0.98 | 0.50 | 1.89 | 0.61 |
| California | 92.22 | 86.46 | 0.60 | 0.55 | 0.43 | 0.38 | 0.48 | 0.75 | 0.78 | 0.43 | 1.62 | 0.57 |
| Brazil | 91.65 | 93.88 | 0.53 | 0.77 | 0.36 | 0.62 | 0.36 | 0.82 | 1.00 | 0.72 | 2.78 | 0.88 |
| Spain | 76.34 | 88.11 | 0.50 | 0.80 | 0.34 | 0.67 | 0.34 | 0.76 | 0.97 | 0.85 | 2.88 | 1.12 |
| Vietnam | 86.86 | 82.17 | 0.44 | 0.58 | 0.28 | 0.41 | 0.28 | 0.77 | 0.97 | 0.46 | 3.43 | 0.60 |
| Sri Lanka | 94.57 | 91.03 | 0.39 | 0.48 | 0.24 | 0.32 | 0.25 | 0.59 | 0.93 | 0.40 | 3.77 | 0.68 |
| Nigeria | 84.34 | 88.06 | 0.38 | 0.77 | 0.23 | 0.62 | 0.23 | 0.98 | 0.98 | 0.63 | 4.26 | 0.65 |
| UAE | 89.93 | 93.13 | 0.33 | 0.60 | 0.20 | 0.43 | 0.23 | 0.51 | 0.61 | 0.74 | 2.70 | 1.47 |
| Myanmar | 93.3 | 90.8 | 0.24 | 0.44 | 0.13 | 0.28 | 0.14 | 0.60 | 0.69 | 0.34 | 4.92 | 0.57 |
| Mozambique | 81.31 | - | 0.84 | - | 0.72 | - | 0.75 | - | 0.96 | - | 1.28 | - |
| China | 93.82 | - | 0.81 | - | 0.68 | - | 0.74 | - | 0.89 | - | 1.21 | - |
| India | 90.22 | - | 0.77 | - | 0.63 | - | 0.65 | - | 0.95 | - | 1.47 | - |
| Philippines | 78.9 | - | 0.76 | - | 0.61 | - | 0.62 | - | 0.99 | - | 1.60 | - |
| Madagascar | 93.56 | - | 0.66 | - | 0.49 | - | 0.52 | - | 0.89 | - | 1.69 | - |
| Uzbekistan | 89.14 | - | 0.58 | - | 0.41 | - | 0.41 | - | 0.99 | - | 2.41 | - |
| Texas | 94.34 | - | 0.57 | - | 0.40 | - | 0.44 | - | 0.79 | - | 1.77 | - |

***Table 4*** *Impact of physics-based postprocessing on flood extent, showing original (pre-refinement) and final (post-refinement) flood area per event, sorted by reduction percentage (descending). Area reduction results from sequential removal of terrain shadow, aerosol artifacts, vegetation false positives, and permanent water misclassifications.*

| Date | Event | Original Flood (km²) | Final Flood (km²) | Area Reduction (km²) | Reduction (%) |
|---|---|---|---|---|---|
| 2023-05-23 | Italy | 76.5 | 62.02 | 14.48 | 18.93 |
| 2022-01-30 | Madagascar | 135.3 | 124.92 | 10.38 | 7.67 |
| 2021-07-26 | China | 133.01 | 124.84 | 8.17 | 6.14 |
| 2017-08-30 | Texas | 1039.15 | 978.3 | 60.85 | 5.86 |
| 2019-09-23 | India | 1418.49 | 1376.16 | 42.33 | 2.98 |
| 2024-05-06 | Brazil | 3075.86 | 2999.48 | 76.38 | 2.48 |
| 2020-11-13 | Philippines | 390.27 | 380.86 | 9.41 | 2.41 |
| 2024-09-18 | Poland | 95.61 | 93.59 | 2.02 | 2.11 |
| 2023-09-10 | Greece | 489.8 | 479.76 | 10.04 | 2.05 |
| 2025-11-07 | Vietnam | 181.71 | 178.47 | 3.24 | 1.78 |
| 2023-01-22 | California | 1913.13 | 1882.6 | 30.53 | 1.6 |
| 2020-05-04 | Uzbekistan | 317.45 | 313.19 | 4.26 | 1.34 |
| 2019-03-22 | Mozambique | 356.03 | 352.46 | 3.57 | 1 |
| 2025-11-30 | Sri Lanka | 49.7 | 49.3 | 0.4 | 0.8 |
| 2024-09-12 | Nigeria | 170.62 | 169.73 | 0.89 | 0.52 |
| 2024-04-19 | UAE | 371.12 | 370.75 | 0.37 | 0.1 |
| 2024-09-20 | Myanmar | 199.79 | 199.7 | 0.09 | 0.04 |
| 2023-09-15 | Libya | 23.32 | 23.32 | 0 | 0 |
| 2024-10-30 | Spain | 141.79 | 141.79 | 0 | 0 |

## Competing interests

The authors declare no competing interests.

## Data and Code Availability

All data generated or analyzed during this study are included in this manuscript. The Prithvi-EO-2.0 model is available at: https://huggingface.co/ibmnasa-geospatial/Prithvi-EO-2.0. The code is available at: https://github.com/NASA-IMPACT/Prithvi-EO-2.0 for fine-tuning and inference.